\documentclass{article}
\usepackage{iclr2027_conference,times}
\usepackage{graphicx}
\usepackage{booktabs}
\usepackage{multirow}
\usepackage{tabularx}
\usepackage{array}
\usepackage{amsmath,amssymb,mathtools}
\usepackage{xspace}
\usepackage{enumitem}
\usepackage[section]{placeins}
\usepackage{float}
\usepackage{caption}
\usepackage{microtype}
\usepackage{url}
\usepackage{hyperref}
\hypersetup{
  hidelinks,
  pdfauthor={Rujie Wu; Haozhe Zhao; Hai Ci; Yizhou Wang},
  pdftitle={Less Data, Faster Convergence: Goal-Driven Data Optimization for Multimodal Instruction Tuning}
}

\graphicspath{{figs/}}
\newcolumntype{Y}{>{\centering\arraybackslash}X}
\newcolumntype{L}[1]{>{\raggedright\arraybackslash}p{#1}}
\setlist[itemize]{leftmargin=1.25em}
\newcommand{\gdo}{\textsc{GDO}\xspace}
\newcommand{\uni}{\textsc{Uni-10x}\xspace}
\newcommand{\vmm}{\textsc{VideoMME}\xspace}
\newcommand{\mvb}{\textsc{MVBench}\xspace}
\newcommand{\lvb}{\textsc{LVBench}\xspace}
\newcommand{\mlvu}{\textsc{MLVU}\xspace}
\newcommand{\tablefontsetup}{\footnotesize\setlength{\tabcolsep}{4pt}}
\definecolor{DeltaNeg}{HTML}{B23A48}
\definecolor{DeltaPos}{HTML}{0F766E}
\newcommand{\dnumstyle}[1]{{\itshape #1}}
\newcommand{\dneg}[1]{{\scriptsize\textcolor{DeltaNeg}{(\dnumstyle{#1})}}}
\newcommand{\dpos}[1]{{\scriptsize\textcolor{DeltaPos}{(\dnumstyle{#1})}}}

\newcommand{\dposplain}[1]{{\scriptsize\textcolor{DeltaPos}{\dnumstyle{#1}}}}
\newcommand{\dnegfull}[1]{{\textcolor{DeltaNeg}{\dnumstyle{#1}}}}
\newcommand{\dposfull}[1]{{\textcolor{DeltaPos}{\dnumstyle{#1}}}}

\title{Less Data, Faster Convergence: Goal-Driven Data Optimization for Multimodal Instruction Tuning}
\author{
Rujie Wu$^{1,*}$ \quad Haozhe Zhao$^{2,*}$ \quad Hai Ci$^{3,\dagger}$ \quad Yizhou Wang$^{1,4,5}$\\
\normalfont\small $^1$School of Computer Science, Peking University, Beijing, China\\
\normalfont\small $^2$Department of Computer Science, University of Illinois Urbana--Champaign,\\
\normalfont\small Urbana, IL, USA\\
\normalfont\small $^3$Show Lab, National University of Singapore, Singapore\\
\normalfont\small $^4$Institute for Artificial Intelligence, Peking University, Beijing, China\\
\normalfont\small $^5$State Key Laboratory of General Artificial Intelligence, Peking University, Beijing, China\\
\normalfont\small $^*$Equal contribution. $^\dagger$Corresponding author.
}

\iclrfinalcopy

\begin{document}
\raggedbottom
\maketitle
\lhead{Preprint}

\begin{abstract}
Multimodal instruction tuning can be compute-inefficient when a fixed training budget is spread across large image--video pools whose examples vary substantially in utility. We present Goal-Driven Data Optimization (\gdo), a framework that computes six sample descriptors for each candidate and constructs optimized 1$\times$ training subsets for different goals. Under a fixed one-epoch Qwen3-VL-8B-Instruct training and evaluation protocol on 32 H20 GPUs, \gdo reaches the benchmark-specific \uni references with fewer training examples and attains higher endpoint accuracy. Each reference is the lowest-scoring endpoint among four historical 512k-sample uniform branches, making the comparison descriptive rather than a matched causal estimate. \gdo reaches these references after 35.4k samples on \mvb, 26.6k on \vmm, 27.3k on \mlvu, and 34.7k on \lvb, while improving endpoint accuracy by 1.38, 1.67, 3.08, and 0.84 percentage points, respectively. The gains are largest on \mvb and \mlvu and smaller on \lvb, consistent with weaker alignment between its ultra-long-video setting and the short-video and image-dominant training pool. The four profiles span low-loss, coverage-oriented, and temporally focused allocations; among them, Temp+ gives the strongest overall endpoint profile.
\end{abstract}

\section{Introduction}
\label{sec:intro}

Modern multimodal assistants have moved from broad visual--language representation learning to instruction-following interaction. Systems such as CLIP~\cite{radford2021clip}, Flamingo~\cite{alayrac2022flamingo}, BLIP-2~\cite{li2023blip2}, PaLI~\cite{chen2022pali}, KOSMOS-1~\cite{huang2023kosmos1}, mPLUG-Owl~\cite{ye2023mplugowl}, LLaVA~\cite{liu2023llava}, InstructBLIP~\cite{dai2023instructblip}, MiniGPT-4~\cite{zhu2023minigpt4}, Qwen-VL~\cite{bai2023qwenvl}, Qwen2-VL~\cite{qwen2vl2024}, InternVL~\cite{chen2024internvl}, and GPT-4V~\cite{team2023gpt4v} have established instruction following as a standard multimodal modeling paradigm. With capable backbones now widely available, a practical question is how to allocate a fixed supervision budget among examples with unequal training value.

This question is sharper in video-centric post-training. Video-capable families such as VideoChatGPT~\cite{maaz2023videochatgpt}, Video-LLaVA~\cite{lin2023videollava}, LLaVA-Video~\cite{zhang2024llavavideo}, LongVA~\cite{zhang2024longva}, MovieChat~\cite{song2024moviechat}, and LongViTU~\cite{wu2025longvitu}, together with large corpora such as Panda-70M~\cite{he2024panda70m}, VISTA~\cite{zhang2024vista400k}, and FineVideo~\cite{finevideo2024}, expand the available video--language supervision. Yet evaluations on \mvb~\cite{li2024mvbench}, \vmm~\cite{fu2024videomme}, \mlvu~\cite{zhou2025mlvu}, and \lvb~\cite{wang2025lvbench} continue to reveal uneven performance across motion, ordering, and long-context reasoning tasks. Related work on open-world visual reasoning likewise shows that free-form concept discrimination remains challenging~\cite{wu2023bongardopenworld}.

This creates a post-training allocation problem. Mixed image--video instruction pools are large, redundant, and heterogeneous, and the value of an additional training example is not uniform. Some samples repeat already-solved patterns; others supply temporal evidence, ordering pressure, or reliable supervision that changes downstream behavior. Prior work on instruction tuning and alignment also shows that curation quality, diversity, and task alignment can matter as much as dataset size in the evaluated settings~\cite{taori2023alpaca,zhou2023lima,dubois2023alpagasus,wang2023m3it}. Scaling-law work~\cite{kaplan2020scaling,hoffmann2022chinchilla,sorscher2023data}, data-quality studies~\cite{wenzek2020ccnet,abbas2023semdedup,gadre2024datacomp,marion2023data}, and instruction-data selection pipelines~\cite{xia2024less,zhang2023activeinstruction} reinforce this point. We ask whether goal-driven data optimization can reduce the supervision needed for multimodal instruction tuning while accelerating convergence under a fixed training protocol.

Existing methods address complementary aspects of data efficiency, but they do not by themselves provide the fixed comparison used in this study. Curriculum and pacing methods~\cite{bengio2009curriculum,kumar2010selfpaced,graves2017automated,fan2018learningtoteach}, sample reweighting~\cite{jiang2018mentornet,ren2018learning}, active selection~\cite{settles2009active,sener2018coreset,kirsch2019batchbald}, valuation~\cite{ghorbani2019data,jia2019data,kwon2023datavaluation}, and instruction-data selection pipelines~\cite{moore2010intelligent,xia2024less,zhang2023activeinstruction} all show that data choice matters. We instead isolate data allocation from changes in model, optimizer, and evaluation protocol.

We answer this question with Goal-Driven Data Optimization (\gdo). \gdo changes only the selected training subset while holding the model, optimizer, checkpoint schedule, and evaluation protocol fixed. It computes six descriptors for each candidate and constructs optimized 1$\times$ subsets subject to explicit budget, modality-mixture, and source-coverage constraints. This design makes observed performance shifts interpretable as data-allocation effects rather than changes in the surrounding recipe.

The four released goal profiles share one descriptor extraction and scoring backbone, and differ primarily in subset size, video-ratio bands, VDS-positive coverage targets, source floors, and oversampling schedules. This design provides a controlled framework for testing how data-optimization goals affect capability, convergence, and sample efficiency under the same training contract.

Two empirical asymmetries motivate this design. First, quantity and usefulness are decoupled in mixed image--video instruction pools. Increasing short-video and image QA by count alone can improve already-strong perception categories while leaving temporal reasoning, ordering, and evidence integration under-served. Second, benchmark alignment is uneven. \mvb and \mlvu organize evaluation around focused subtasks involving motion, ordering, and comparison. By contrast, \lvb emphasizes ultra-long-video understanding and is less closely matched to the short-video and image-dominant training pool used here. This mismatch offers one plausible explanation for the smaller gains observed on \lvb. Thus, \gdo is a controlled data-allocation framework whose effect depends on the match between the candidate pool and target evaluation distribution.

Sample efficiency and convergence must be evaluated together. A small optimized subset is meaningful only if it reaches useful capability earlier without trading final quality for speed. Conversely, a final-score gain is incomplete if it requires a larger training budget. Our evaluation therefore freezes one \uni reference per benchmark and asks whether goal-driven data optimization can reach or exceed that reference earlier while attaining stronger reported endpoints.

\begin{figure*}[t]
  \centering
  \includegraphics[width=\linewidth]{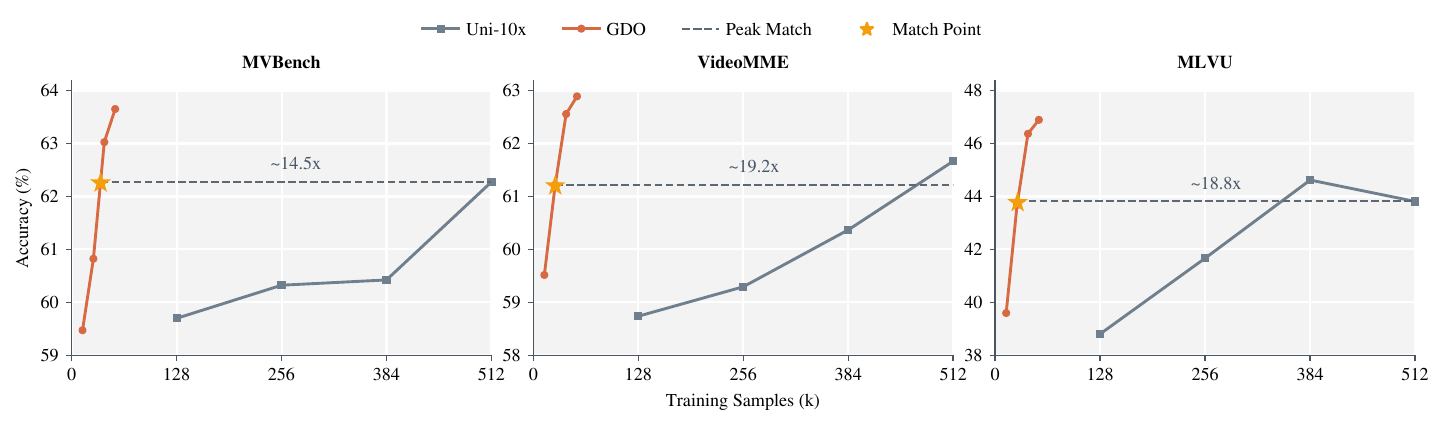}
  \caption{\textbf{Peak Match with Less Data.} Accuracy is plotted against training samples for \mvb, \vmm, and \mlvu. The dashed line marks the fixed \uni reference, and the star marks its first crossing under linear interpolation between adjacent displayed \gdo checkpoints. Table~\ref{tab:intro-snapshot} reports the corresponding Peak Match and Reduction values.}
  \label{fig:teaser}
\end{figure*}

Figure~\ref{fig:teaser} shows that a smaller optimized subset can reach and exceed the fixed \uni reference earlier in training. Table~\ref{tab:intro-snapshot} reports more than 10$\times$ peak-match data reduction on all four benchmarks. The gains vary across benchmarks: they are largest on \mvb and \mlvu and smaller on \lvb. This pattern is consistent with closer alignment between the available pool and the focused subtasks in \mvb and \mlvu, and weaker alignment with the ultra-long-video regime of \lvb.

\begin{table*}[!t]
\centering
\caption{\textbf{Peak Match and Reduction.} Peak Match is the descriptive first crossing of the fixed \uni reference under linear interpolation between adjacent displayed checkpoints; it is not an independently evaluated checkpoint. Reduction is relative to the 512k-sample \uni budget. Scores are accuracy (\%), and $\Delta$ is the improvement over \uni in percentage points (pp).}
\label{tab:intro-snapshot}
\tablefontsetup
\begin{tabularx}{\textwidth}{lYYYYY}
\toprule
Benchmark & \uni & \gdo & $\Delta$ (pp) & Peak Match & Reduction \\
\midrule
\mvb & 62.27 & 63.65 & \dposfull{+1.38} & 35.4k & 14.5$\times$ \\
\vmm & 61.22 & 62.89 & \dposfull{+1.67} & 26.6k & 19.2$\times$ \\
\mlvu & 43.81 & 46.89 & \dposfull{+3.08} & 27.3k & 18.8$\times$ \\
\lvb & 40.22 & 41.06 & \dposfull{+0.84} & 34.7k & 14.8$\times$ \\
\bottomrule
\end{tabularx}
\end{table*}

\begin{samepage}
\noindent Taken together, the results support three claims:
\begin{itemize}[leftmargin=*,topsep=0.25em,itemsep=0.15em,label=\raisebox{0.2ex}{\footnotesize$\bullet$}]
  \item \textbf{Less Data}. Optimized 1$\times$ subsets can outperform the fixed \uni baseline with far fewer training samples.
  \item \textbf{Faster Convergence}. The gains appear as earlier frontier crossings, not only as stronger reported endpoints.
  \item \textbf{Goal-Driven Data Optimization}. Different allocation goals produce distinct capability profiles, with Temp+ giving the strongest overall result among the tested profiles.
\end{itemize}
\end{samepage}

\noindent Our contributions are therefore threefold:
\begin{itemize}[leftmargin=*,label=\raisebox{0.2ex}{\footnotesize$\bullet$}]
  \item We formulate multimodal instruction-data allocation under fixed training and evaluation settings as a goal-driven data optimization problem.
  \item We instantiate \gdo as a reproducible pipeline that combines six sample descriptors, shared candidate scoring, and goal-specific optimization presets over budget, video composition, coverage, and source balancing.
  \item We provide benchmark, trajectory, subtask, and ablation evidence that the selected subsets reach the fixed reference earlier and attain higher endpoint accuracy under this contract, while documenting how the gains vary across benchmarks.
\end{itemize}

\section{Method}
\label{sec:method}

\subsection{Problem Setup}

Let $\mathcal{D}=\{x_i\}_{i=1}^{N}$ denote the shared multimodal instruction pool, $g$ a data-allocation goal, and $b$ an evaluation benchmark. For each goal, \gdo builds a 1$\times$ optimized subset $S_g$. For each benchmark, $U_b$ denotes the frozen endpoint score of the weakest 512k-sample uniform branch among the four strategy suites. Thus $U_b$ is a benchmark-specific reporting reference, not one shared uniform run. The backbone, one-epoch SFT recipe, and evaluation implementation remain fixed across all branches.

\begin{figure*}[t]
\centering
\includegraphics[width=\textwidth]{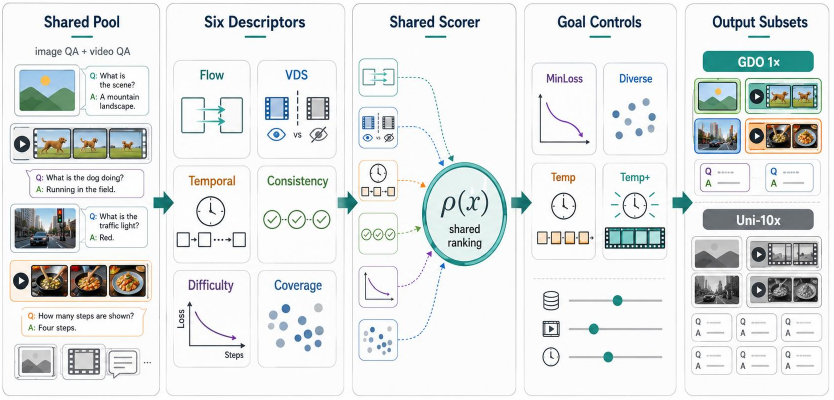}
\caption{\textbf{Subset Construction.} \gdo computes six sample descriptors over one shared pool, applies a shared score and goal-specific feasibility presets, and trains the resulting 1$\times$ subsets under the same backbone and SFT recipe. Evaluation uses the frozen benchmark-specific \uni references defined in Section~\ref{sec:exp}.}
\label{fig:method}
\end{figure*}

The released pipeline is deterministic and staged. It first computes six sample descriptors, then maps each candidate to a shared score $\rho(x)$, and finally applies a goal-specific feasibility preset $\mathcal{C}_g$ that controls budget, video ratio, temporal-positive coverage, source floors, and candidate oversampling as follows:
\begin{equation}
S_g = \mathrm{Build}\!\left(\mathcal{D}; \rho, \mathcal{C}_g\right), \qquad
\Delta_{g,b} = \mathrm{Eval}_b(S_g)-U_b.
\label{eq:gdo}
\end{equation}
At the paper level, the intervention is restricted to subset construction. As summarized in Figure~\ref{fig:method}, the descriptors provide comparable cues, the shared score defines preference, and the goal preset defines admissibility. Subset construction is benchmark-blind: benchmark test identities and answers never enter descriptor extraction, scoring, or filtering. The same descriptor table is reused across profiles. Benchmark-wise selection of $U_b$ affects the reporting reference, not subset construction, and is disclosed throughout.

\subsection{Six Sample Descriptors}

Each candidate is represented by a six-dimensional descriptor vector
\begin{equation}
\mathbf{m}(x)=\left[m_{\mathrm{flow}},m_{\mathrm{vds}},m_{\mathrm{tnc}},m_{\mathrm{sc}},m_{\mathrm{ppl}},m_{\mathrm{cov}}\right].
\end{equation}
All six descriptors are computed from candidate training examples with frozen probes or deterministic statistics; none uses benchmark labels, test examples, or downstream rewards.

For a sample $x=(v_{1},\ldots,v_{T},q,y)$, \gdo instantiates the six descriptors as follows:

\noindent\textbf{Flow} measures average optical-flow magnitude across adjacent frames
\begin{equation}
m_{\mathrm{flow}}(x)=\frac{1}{T-1}\sum_{t=1}^{T-1}\frac{1}{|\Omega|}\sum_{p\in\Omega}\|u_t(p)\|_2,
\end{equation}
where $u_t$ is the dense flow field between frames $v_t$ and $v_{t+1}$. We average per-pixel motion magnitude over adjacent frame pairs; larger values therefore correspond to stronger visible motion cues.

\noindent\textbf{VDS (video-dependence score)} uses the blind-vs-video loss gap
\begin{equation}
m_{\mathrm{vds}}(x)=\mathcal{L}_{\mathrm{blind}}(x)-\mathcal{L}_{\mathrm{video}}(x),
\end{equation}
where $\mathcal{L}_{\mathrm{blind}}$ and $\mathcal{L}_{\mathrm{video}}$ are computed with the same frozen Qwen3-VL-8B-Instruct probe under blind and video-conditioned inputs. We evaluate each question--answer pair with and without video frames; a larger VDS means that visual input reduces the answer loss more strongly.

\noindent\textbf{Temporal necessity} assigns a question-level proxy score
\begin{equation}
m_{\mathrm{tnc}}(x)=T(q),
\end{equation}
where $T(q)\in[0,1]$ is a temporal-necessity judgment from the frozen probe. We map its judgment to a scalar proxy; higher values indicate that the question is more likely to require ordering, change, duration, or cross-frame reasoning rather than static visual lookup.

\noindent\textbf{Self-consistency} measures agreement across stochastic decodes
\begin{equation}
m_{\mathrm{sc}}(x)=\operatorname{mean}_{i<j}\,\mathrm{Jac}(a_i,a_j),
\end{equation}
where $a_i$ are stochastic decodes. We compute mean pairwise Jaccard similarity across repeated sampled answers; higher agreement indicates a more stable probe response.

\noindent\textbf{PPL-like difficulty} uses the exponentiated teacher-forced video loss
\begin{equation}
m_{\mathrm{ppl}}(x)=\exp(\mathcal{L}_{\mathrm{video}}(x)).
\end{equation}
We exponentiate the teacher-forced video-conditioned loss of the target answer; larger values indicate that the frozen probe finds the sample harder to fit.

\noindent\textbf{Coverage} combines semantic clustering and source statistics
\begin{equation}
m_{\mathrm{cov}}(x)=C(x;\mathcal{N}_{\mathrm{text}},\mathcal{N}_{\mathrm{vision}},s(x)),
\end{equation}
where $C(\cdot)$ combines the local text neighborhood $\mathcal{N}_{\mathrm{text}}$, visual neighborhood $\mathcal{N}_{\mathrm{vision}}$, and source identity $s(x)$. Larger values favor semantic and source-level breadth over already over-represented regions of the pool.

The builder consumes these descriptors in two complementary forms, with their normalized combinations contributing to a merged quality summary used by the shared scorer while the raw descriptors remain available to feasibility controls and ablations. As a result, one descriptor layer supports both global ranking and goal-specific composition constraints, while still covering motion content, video dependence, temporal demand, answer stability, optimization difficulty, and semantic/source breadth. \gdo therefore preserves multiple allocation cues that would otherwise be compressed into a single quality score.

\subsection{Scoring and Feasibility}

The descriptor table is used in two coupled but distinct ways. A shared score $\rho(x)$ determines the candidate ranking, while the goal-specific preset $\mathcal{C}_g$ determines which ranked candidates are admissible under a given allocation target. The four profiles share one scoring backbone and differ mainly in budget and composition constraints. For short-video samples, the released score is
\begin{equation}
\rho_{\mathrm{vid}}(x)=0.35\,\tanh\!\left(\frac{b_{\mathrm{vid}}(x)}{3}\right)
+0.95\,z_{\mathrm{vds3}}(x)+0.35\,z_{\mathrm{qual}}(x),
\label{eq:score}
\end{equation}
where $z_{\mathrm{vds3}}$ is a normalized video-dependence term derived from $\mathcal{L}_{\mathrm{video}}$, $\mathcal{L}_{\mathrm{blind}}$, and frame diversity, and $z_{\mathrm{qual}}$ is the normalized merged \texttt{quality\_score}. The base term is
\begin{equation}
b_{\mathrm{vid}}(x)=q_{\mathrm{text}}(x)+0.85\,d(x)+0.9\,a(x)+0.55\,t(x)+0.15\,r_{\mathrm{src}}(x)
\end{equation}
where $q_{\mathrm{text}}$ is a heuristic text-quality prior, $d$ is a medium-difficulty preference, $a$ is a bin-alignment term over duration, temporal, question-length, and source strata, $t$ is a temporal bonus, and $r_{\mathrm{src}}$ is a source-rarity term.

For image QA, the released builder uses the same logic without the video-only terms
\begin{equation}
\rho_{\mathrm{img}}(x)=0.90\,\tanh\!\left(\frac{b_{\mathrm{img}}(x)}{3}\right)+0.15\,z_{\mathrm{qual}}(x),
\end{equation}
with $b_{\mathrm{img}}(x)$ omitting the temporal bonus and using a slightly stronger text-quality weight. Because the contributing terms are normalized before mixing, these coefficients should be read as fixed mixture weights in the released builder rather than as learned parameters. They are set once to balance the base heuristic, video-dependence, and merged-quality contributions, and are then shared across all four goal profiles rather than retuned per goal or benchmark. The ablations later in the paper therefore test component importance under one fixed scorer instead of re-optimizing these weights.

Because $\rho$ is shared, changing $g$ primarily changes admissibility. A global sort by $\rho(x)$ would over-select abundant, linguistically clean image QA and under-select rarer temporal video samples. The feasibility controls instead enforce modality mix, temporal coverage, and source breadth alongside score. The same descriptor table and scorer can therefore support multiple allocation goals by changing only the feasibility constraints.

Operationally, the builder maintains per-stratum top-$k$ reservoirs and fills the subset in stages that include temporal-video minima, video-ratio minima, source floors, stratum quotas, and finally the global score tail with reservoir fallback. Deduplication and QA-per-video caps are enforced throughout. This staged fill is the released realization of Eq.~\eqref{eq:gdo}.

\subsection{Goal Profiles}

The paper reports four released goal profiles within \gdo, each corresponding to a different setting of the same builder with shared scoring and different feasibility presets. For paper reporting, all four profiles are compared against the frozen benchmark-specific \uni references:

\noindent\textbf{MinLoss} explicitly targets the lowest training loss during subset construction. It therefore favors the easiest-to-fit supervision under the smallest budget (12.9k) and applies the lightest temporal and video pressure.

\noindent\textbf{Diverse} shifts the objective from minimum loss to broader semantic and source coverage. Its larger subset (42.9k) and stronger source floors make it the most conservative profile against collapse onto only the easiest samples.

\noindent\textbf{Temp} optimizes for stronger temporal usefulness. It allocates 33.3k samples while raising the target video ratio and temporal-positive coverage so that more of the selected budget is spent on temporally informative video supervision.

\noindent\textbf{Temp+} pushes that temporal objective further. It increases both subset size and temporal pressure (53.3k, 58.7\% selected video), making it the strongest temporal profile in the released suite.

Together, the profiles span low-loss, coverage-oriented, and temporally focused allocations. They share $\rho$ and differ through the controls in $\mathcal{C}_g$, allowing the resulting frontier shifts to be compared without changing the scorer.

\begin{table*}[t]
\centering
\caption{\textbf{Overall Results.} Rows report the four released 1$\times$ \gdo profiles. Each cell gives accuracy (\%), with the pp difference relative to the frozen benchmark-specific 512k-sample \uni reference in parentheses.}
\label{tab:main-results}
\tablefontsetup
\setlength{\tabcolsep}{3pt}
\begin{tabularx}{\textwidth}{L{0.14\textwidth}L{0.10\textwidth}YYYY}
\toprule
Setting & 1$\times$ samples & MVBench & VideoMME & MLVU & LVBench \\
\midrule
MinLoss & 12.9k & \mbox{63.63 \dpos{+1.35}} & \mbox{62.30 \dpos{+1.07}} & \mbox{45.84 \dpos{+2.03}} & \mbox{38.86 \dneg{-1.36}} \\
Diverse & 42.9k & \mbox{63.12 \dpos{+0.85}} & \mbox{61.33 \dpos{+0.11}} & \mbox{46.05 \dpos{+2.24}} & \mbox{39.90 \dneg{-0.32}} \\
Temp & 33.3k & \mbox{62.05 \dneg{-0.23}} & \mbox{62.04 \dpos{+0.81}} & \mbox{45.26 \dpos{+1.45}} & \mbox{40.28 \dpos{+0.06}} \\
Temp+ & 53.3k & \mbox{\textbf{63.65} \dpos{+1.38}} & \mbox{\textbf{62.89} \dpos{+1.67}} & \mbox{\textbf{46.89} \dpos{+3.08}} & \mbox{\textbf{41.06} \dpos{+0.84}} \\
\bottomrule
\end{tabularx}
\end{table*}

\begin{figure}[H]
  \centering
  \includegraphics[width=0.72\linewidth]{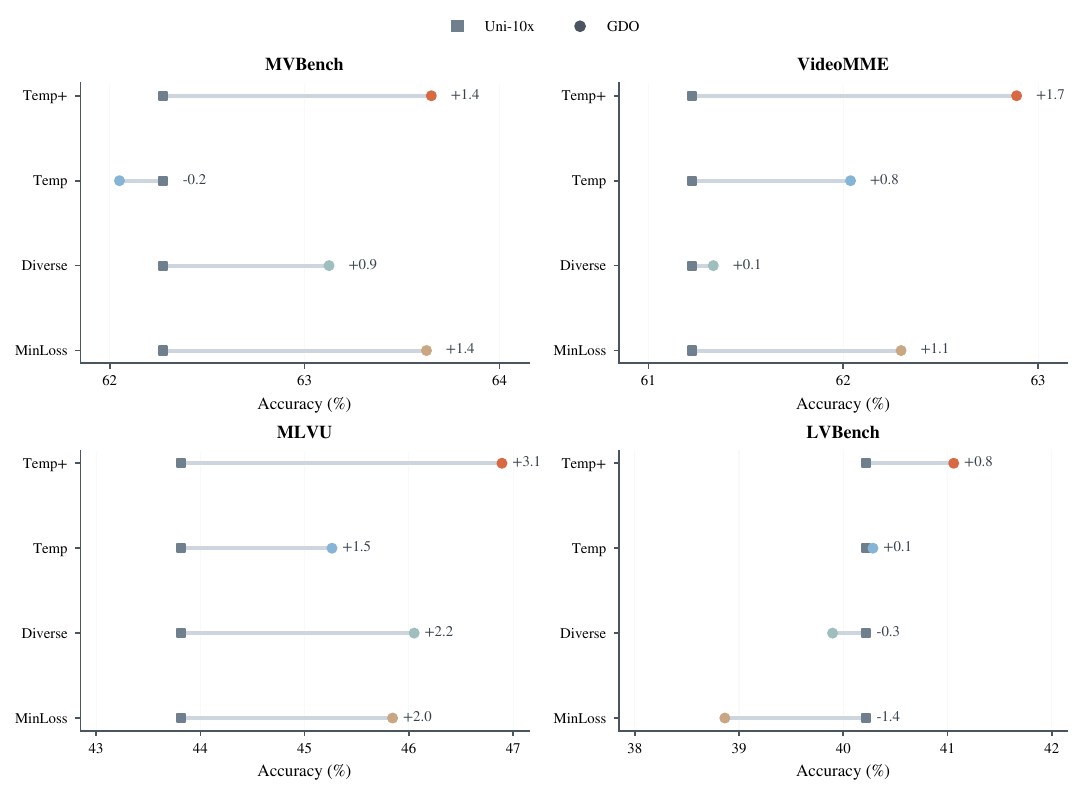}
  \caption{\textbf{Frontier Shifts by Goal.} For each benchmark, dots mark the strongest operating points of the four \gdo profiles, while the square marks the frozen benchmark-specific \uni reference. These are descriptive comparisons, not matched causal contrasts.}
  \label{fig:bestof4-comparison}
\end{figure}

\section{Experiments}
\label{sec:exp}

\subsection{Experimental Setup}

\noindent\textbf{Data.}
We build all subsets from one shared pool consisting of the full LLaVA-OneVision image-QA data~\cite{li2024llavaonevision} and the full LLaVA-Video short-video QA data~\cite{zhang2024llavavideo}. Each \gdo profile constructs one optimized 1$\times$ subset from that pool.

\noindent\textbf{Training.}
We follow one fixed one-epoch Qwen3-VL-8B-Instruct SFT~\cite{qwen3vl2025} recipe on 32 H20 GPUs with batch size 1 per GPU and global batch size 32. Only the selected subset and its resulting sample count vary across \gdo profiles.

\noindent\textbf{Comparison and trajectories.}
For each benchmark, we freeze the lowest-scoring endpoint among four historical 512k-sample \uni branches as $U_b$ and use it for every reported delta. Consequently, the comparisons are descriptive and are not matched causal controls. For trajectory figures, four points are retained from five saved checkpoints using a fixed screening rule that removes one noisy point and preserves the most monotone subsequence. Their measured scores are unchanged, but their horizontal positions are pseudo-quarter sample counts based on each branch's maximum paper-facing budget. Endpoint scores are direct checkpoint measurements; intermediate positions and peak-match crossings are descriptive visual summaries.

\noindent\textbf{Benchmarks.}
We evaluate \mvb~\cite{li2024mvbench}, \vmm~\cite{fu2024videomme}, \mlvu~\cite{zhou2025mlvu}, and \lvb~\cite{wang2025lvbench}. \mvb and \mlvu are subtask-oriented, \vmm is broader and less tightly factorized, and \lvb emphasizes ultra-long-video understanding. This combination tests whether the same data-allocation strategy transfers across focused temporal subtasks, broader video QA, and a substantially longer-video regime. All scores are reported as accuracy (\%), with differences in percentage points (pp).

\subsection{Benchmark Results}

Table~\ref{tab:main-results} gives the main benchmark result. With fewer than 512k training samples, \gdo improves \mvb, \vmm, \mlvu, and \lvb by 1.38, 1.67, 3.08, and 0.84 pp, respectively. The largest gains occur on \mvb and \mlvu, whose focused subtasks emphasize motion, order, and temporal evidence. This association is consistent with the intended allocation, although the experiment does not isolate benchmark structure as its cause. \vmm also improves, showing that the gains are not confined to one subtask-heavy evaluation style.

Figure~\ref{fig:bestof4-comparison} shows the same result from the convergence side. Linear interpolation between adjacent displayed checkpoints places the first reference crossing at 35.4k samples on \mvb, 26.6k on \vmm, 27.3k on \mlvu, and 34.7k on \lvb. These descriptive peak-match points operationalize ``less data, faster convergence'' at the recorded checkpoint cadence; they are not independently evaluated checkpoints. Together, Table~\ref{tab:main-results} and Figure~\ref{fig:bestof4-comparison} show changes in both endpoint accuracy and the observed training trajectory.

Looking across the four profiles is more informative than focusing only on the headline \gdo result. MinLoss targets the lowest-loss supervision and therefore reaches useful operating points under the smallest budget. It isolates what low-loss allocation alone provides, namely strong sample efficiency with limited emphasis on the harder temporal evidence required by the most demanding temporal tasks. Diverse behaves differently. It restores broader semantic and source coverage, which keeps it close to the fixed \uni reference on several benchmarks and avoids collapsing onto only the easiest supervision, but it remains conservative on temporal specialization.

Temp and Temp+ allocate more of the budget to temporally informative video supervision. The intermediate profiles are not strictly ordered on every benchmark, but Temp+ gives the strongest endpoint result on all four. The profile comparison therefore supports a difference among allocation goals without implying a monotonic response at every intermediate setting.

The \lvb result clarifies the scope of the method. It evaluates ultra-long videos, often much longer than the clips represented in LLaVA-Video, while the training pool also contains substantial image QA from LLaVA-OneVision. Within the tested profiles, \lvb changes from negative or near-zero differences for MinLoss and Diverse to positive differences for Temp and Temp+. This pattern is consistent with a benefit from temporal allocation, while the available pool limits the size of the observed gain.

\subsection{Subtask Analysis}

\begin{table*}[t]
\centering
\caption{\textbf{Subtask Deltas.} Values are subtask deltas in pp relative to the frozen benchmark-specific \uni reference. Unlike Table~\ref{tab:positive-subtasks}, which highlights the largest temporal wins, this table keeps adverse subtasks visible.}
\label{tab:subtask-deltas}
\tablefontsetup
\setlength{\tabcolsep}{3pt}
\begin{tabularx}{\textwidth}{L{0.14\textwidth}YYYYYY}
\toprule
Setting & MVB Motion & MVB Reasoning & VMM Temp. Perc. & MLVU Order & MLVU SportsQA & MLVU Ego \\
\midrule
MinLoss & \dnegfull{-1.1} & \dposfull{+3.2} & \dposfull{+9.1} & \dposfull{+5.7} & \dposfull{+11.1} & \dnegfull{-15.1} \\
Diverse & \dposfull{+1.5} & \dnegfull{-2.2} & \dposfull{+10.9} & \dposfull{+10.0} & \dposfull{+8.3} & \dnegfull{-5.7} \\
Temp & \dnegfull{-1.2} & \dposfull{+0.6} & \dposfull{+7.3} & \dposfull{+4.3} & \dposfull{+5.6} & \textcolor{DeltaPos}{\textbf{\dnumstyle{+1.9}}} \\
Temp+ & \textcolor{DeltaPos}{\textbf{\dnumstyle{+0.8}}} & \textcolor{DeltaPos}{\textbf{\dnumstyle{+3.8}}} & \textcolor{DeltaPos}{\textbf{\dnumstyle{+18.2}}} & \textcolor{DeltaPos}{\textbf{\dnumstyle{+7.1}}} & \textcolor{DeltaPos}{\textbf{\dnumstyle{+11.1}}} & \dnegfull{-3.8} \\
\bottomrule
\end{tabularx}
\end{table*}

Table~\ref{tab:subtask-deltas} then shows the same movement over the main subtasks while keeping adverse subtasks visible. Temp+ is strongest because it combines positive motion, positive reasoning, and very large temporal-perception gains without giving up the strongest order-related subtasks. Diverse preserves some motion and order behavior while remaining more coverage-oriented, whereas MinLoss trades isolated wins for larger regressions elsewhere. Temp sits between them and makes the transition from broad data allocation to explicitly temporal allocation visible.

\subsection{Ablation Study}

The profile comparison already shows that stronger temporal goals matter. We next ask whether Temp+ is driven by one dominant score term or by the combination of several cues. Table~\ref{tab:ablation} supports the latter reading.

The ablation pattern is benchmark-dependent rather than collapsing onto one dominant feature. On \vmm, removing VDS or the PPL-like term costs about one point. On \mvb and especially \mlvu, removing self-consistency produces the largest decrease among the single-component ablations, which is consistent with answer stability being a useful cue in this setting. Some single removals help one benchmark while hurting others, further arguing against a single monotone driver. The joint removal is more damaging on \mvb than any individual ablation, suggesting that Temp+ relies on multiple scoring ingredients and feasibility controls.

\subsection{Discussion}

The optimized subsets change both the endpoint and the observed training trajectory: they reach the fixed \uni reference at earlier checkpoints and finish above it under the same backbone, optimizer, and evaluation protocol.

The gains are capability-dependent. \mvb and \mlvu benefit most, while \lvb improves more modestly under the short-video and image-heavy training pool. The result is therefore conditional on the candidate pool: allocation can only emphasize capabilities for which the pool contains relevant supervision.

The released profiles form an allocation spectrum rather than a single operating point. MinLoss is the most budget-efficient setting, Diverse restores broader coverage, and Temp/Temp+ place greater emphasis on temporal supervision. The subtask and ablation results show concentrated gains alongside adverse cells. Under the fixed protocol studied here, data allocation changes the efficiency--capability trade-off, and the appropriate profile depends on the target capability and source distribution.

\FloatBarrier

\section{Related Work}
\label{sec:related}

CLIP and CoCa establish large-scale image--text pretraining~\cite{radford2021clip,yu2022coca}. Flamingo, BLIP-2, PaLI, PaLI-X, KOSMOS-1, and mPLUG-Owl extend this line toward few-shot behavior, generation, and unified multimodal modeling~\cite{alayrac2022flamingo,li2023blip2,chen2022pali,chen2023palix,huang2023kosmos1,ye2023mplugowl}. LLaVA, LLaVA-1.5, LLaVA-NeXT-Interleave, InstructBLIP, and MiniGPT-4 turn such backbones into assistants~\cite{liu2023llava,liu2023llava15,li2024llavanextinterleave,dai2023instructblip,zhu2023minigpt4}, while Qwen-VL, Qwen2-VL, InternVL, InternVL 2.5, GPT-4, and GPT-4V further strengthen this family~\cite{bai2023qwenvl,qwen2vl2024,chen2024internvl,liu2024internvl2,achiam2023gpt4,team2023gpt4v}. These systems motivate strong fixed backbones, but leave open how to allocate a limited multimodal SFT budget once the backbone is fixed.

Alpaca, LIMA, and AlpaGasus study compact or curated instruction sets~\cite{taori2023alpaca,zhou2023lima,dubois2023alpagasus}. M3IT and InstructBLIP extend instruction tuning to heterogeneous multimodal sources~\cite{wang2023m3it,dai2023instructblip}. \gdo combines this motivation with benchmark-specific frozen reporting references; the resulting contrasts are descriptive rather than matched causal estimates.

VideoChatGPT, Video-LLaVA, Chat-UniVi, MovieChat, and LLaVA-Video extend instruction following from images to videos~\cite{maaz2023videochatgpt,lin2023videollava,jin2024chatunivi,song2024moviechat,zhang2024llavavideo}. LongVA, VideoAgent, and LongViTU push toward longer context~\cite{zhang2024longva,fan2024videoagent,wu2025longvitu}, while Panda-70M, VISTA, and FineVideo enlarge the video--language supervision pool~\cite{he2024panda70m,zhang2024vista400k,finevideo2024}. \mvb, \vmm, \mlvu, and \lvb expose uneven performance across temporal subtasks~\cite{li2024mvbench,fu2024videomme,zhou2025mlvu,wang2025lvbench}, making data allocation relevant to video-centric post-training.

Scaling analyses study compute--data balance~\cite{kaplan2020scaling,hoffmann2022chinchilla,sorscher2023data}. Curriculum and pacing methods change when examples are seen~\cite{bengio2009curriculum,kumar2010selfpaced,graves2017automated,fan2018learningtoteach}, while sample reweighting and forgetting analyses study which examples matter most~\cite{jiang2018mentornet,ren2018learning,toneva2018forgetting,paul2021deep}. Active selection and data valuation make that choice explicit~\cite{settles2009active,sener2018coreset,kirsch2019batchbald,ghorbani2019data,jia2019data,kwon2023datavaluation}. Corpus filtering and pre-training data management improve large training pools~\cite{wenzek2020ccnet,abbas2023semdedup,gadre2024datacomp,peng2025dataman,zhuang2025metarater}, while instruction-data selection and refinement target post-training supervision~\cite{marion2023data,moore2010intelligent,xia2024less,zhang2023activeinstruction,zhao2024ultraedit,lyu2025cream,yu2025improving}. These studies motivate data-quality and selection signals. \gdo combines multiple descriptors with goal-specific feasibility constraints. Self-consistency provides a complementary stability cue~\cite{wang2022selfconsistency}.

\section{Conclusion}
\label{sec:conclusion}

We presented \gdo, a goal-driven data optimization framework for multimodal instruction tuning. Under the fixed Qwen3-VL-8B-Instruct protocol, \gdo reaches the frozen benchmark-specific \uni references with fewer samples and delivers higher reported endpoint accuracy on \mvb, \vmm, \mlvu, and \lvb. Because each reference is the weakest historical uniform branch for that benchmark, these comparisons describe the original reporting protocol rather than matched causal effects. The released profiles show distinct efficiency--capability trade-offs.

These results frame multimodal SFT data as an allocation problem as well as a scaling problem. Combining sample descriptors with shared scoring and goal-specific feasibility controls exposes that choice while keeping the remaining training protocol fixed. The conclusion is bounded by the candidate pool: improvements are largest when its supervision is aligned with the target capability, whereas ultra-long-video evaluation remains weakly matched by the data used here.

\section*{Reproducibility Statement}
The main text specifies the candidate pool, subset sizes, backbone, training protocol, checkpoints, evaluation benchmarks, and reported metrics. The appendix provides the shared scoring functions, goal-profile constraints, full trajectories, subtask analyses, ablations, limitations, and qualitative examples. The accompanying code release contains the data-processing, subset-construction, training, evaluation, and figure-generation implementations.

\section*{AI Use Statement}
Generative AI tools were used primarily to check grammar and improve the clarity and readability of the manuscript. The authors reviewed all AI-assisted edits and take responsibility for the final manuscript.

\bibliography{references}
\bibliographystyle{iclr2027_conference}

\clearpage
\appendix
\section*{Appendix}
\section{Additional Analysis}
\label{sec:supp-extended}

\subsection{Trajectories}

Figure~\ref{fig:supp-efficiency} extends the peak-match result to full trajectories on all four benchmarks under the same fixed comparison. The earlier crossing pattern is visible beyond a single checkpoint. The separation is largest on \mvb, \vmm, and \mlvu and smaller on \lvb. The latter pattern is consistent with, but does not by itself establish, a mismatch between the ultra-long evaluation videos and the short-video and image-dominant training pool.

\begin{center}
  \includegraphics[width=0.96\textwidth]{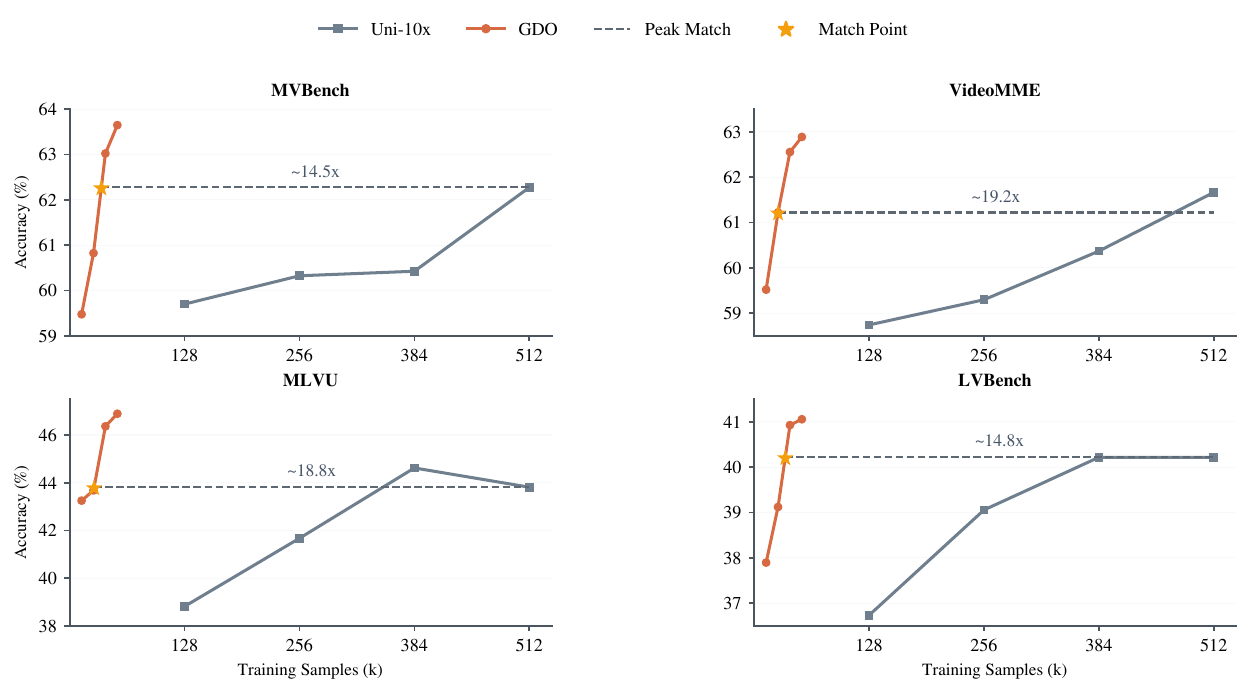}
  \captionof{figure}{\textbf{Trajectories.}
  Each panel compares \gdo with its frozen benchmark-specific 512k-sample \uni reference. Stars denote the first linearly interpolated crossing between adjacent displayed checkpoints.}
  \label{fig:supp-efficiency}
\end{center}
\FloatBarrier

\subsection{Goal Profiles}

Figure~\ref{fig:supp-profile-frontiers} places the four released goal profiles against the same frozen per-benchmark \uni reference. MinLoss occupies the earliest low-budget regime, whereas Diverse stays closer to a broad-coverage allocation. Temp and Temp+ raise both the selected video ratio and the temporal-positive floor; Temp+ attains the strongest endpoint profile, although the intermediate profiles are not strictly ordered on every benchmark.

\begin{center}
  \includegraphics[width=0.96\textwidth]{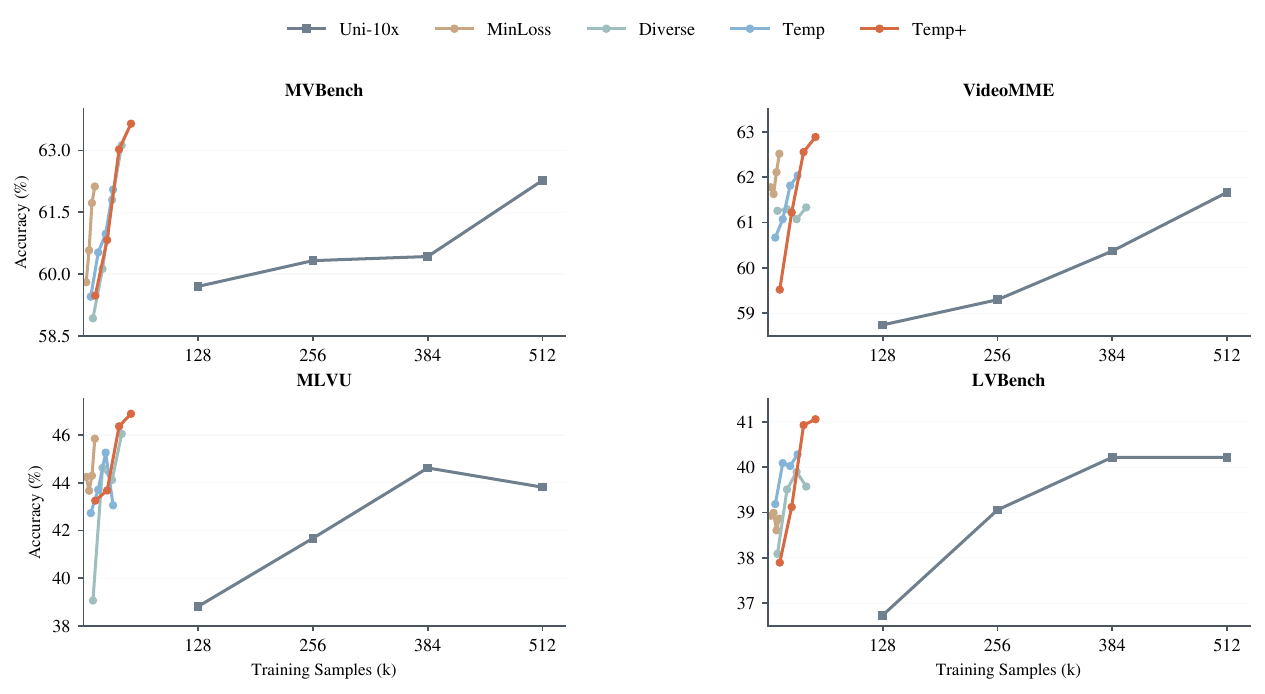}
  \captionof{figure}{\textbf{Goal Profiles.}
  Each panel overlays MinLoss, Diverse, Temp, and Temp+ against the same frozen benchmark-specific \uni reference.}
  \label{fig:supp-profile-frontiers}
\end{center}

\subsection{Ablation}

The profile comparison does not identify which parts of the Temp+ builder contribute most. Figure~\ref{fig:supp-tempplus-ablation} extends the main-paper ablation into a trajectory comparison while keeping the remaining training and evaluation protocol fixed. On \vmm, removing VDS or PPL hurts most. On \mlvu, removing self-consistency causes the largest drop among the tested single-component removals, which is consistent with answer stability being useful in this setting. On \mvb, the joint removal is most damaging, showing that no single ablation accounts for the full change.

\begin{center}
  \includegraphics[width=0.98\textwidth]{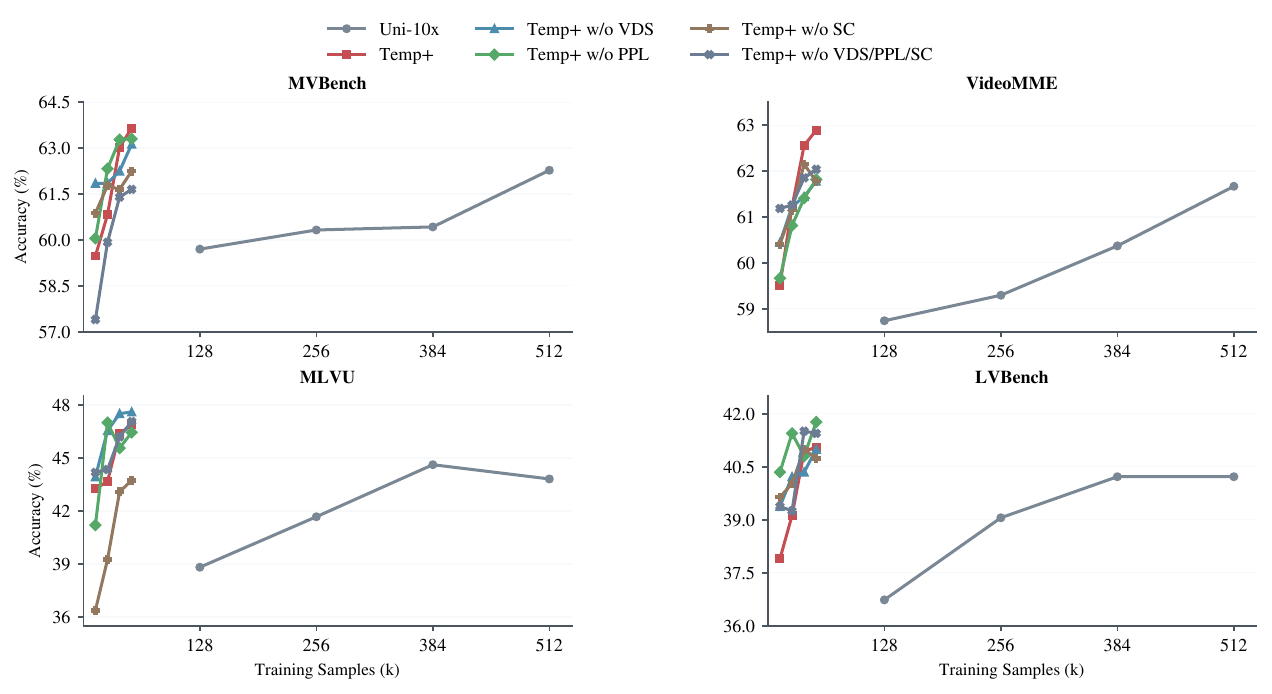}
  \captionof{figure}{\textbf{Ablation Trajectories.}
  Each curve removes one or more Temp+ score contributions while keeping the rest of the train/eval contract fixed.}
  \label{fig:supp-tempplus-ablation}
\end{center}
\FloatBarrier

\section{Subtask Analysis}
\label{sec:supp-subtasks}

Table~\ref{tab:positive-subtasks} reports a post hoc descriptive ranking of the five temporally related subtasks with the largest gain for each profile. Recurring categories include temporal perception in \vmm, moving-count and order-sensitive subtasks in \mvb, and Order and SportsQA in \mlvu. The main-text table and heatmap retain adverse outcomes.

\begin{table}[t]
\centering
\caption{\textbf{Temporal Subtask Gains.} The five temporally related subtasks with the largest gains for each \gdo profile over the fixed per-benchmark \uni reference. Scores are accuracy (\%), and $\Delta$ is the difference in percentage points.}
\label{tab:positive-subtasks}
\small
\begin{tabularx}{\textwidth}{lXccc}
\toprule
Benchmark & Subtask & \uni (\%) & GDO (\%) & $\Delta$ (pp) \\
\midrule
\multicolumn{5}{l}{\textbf{MinLoss}} \\
MLVU & Order & 25.71 & 31.43 & \dposplain{+5.71} \\
MVBench & Character Order & 67.50 & 74.50 & \dposplain{+7.00} \\
VideoMME & Temporal Perception & 67.27 & 76.36 & \dposplain{+9.09} \\
MVBench & Moving Count & 53.00 & 63.00 & \dposplain{+10.00} \\
MLVU & SportsQA & 36.11 & 47.22 & \textbf{\dposplain{+11.11}} \\
\midrule
\multicolumn{5}{l}{\textbf{Diverse}} \\
MLVU & SportsQA & 36.11 & 44.44 & \dposplain{+8.33} \\
MVBench & State Change & 61.00 & 70.50 & \dposplain{+9.50} \\
MLVU & Order & 25.71 & 35.71 & \dposplain{+10.00} \\
VideoMME & Temporal Perception & 67.27 & 78.18 & \dposplain{+10.91} \\
MVBench & Moving Count & 53.00 & 66.00 & \textbf{\dposplain{+13.00}} \\
\midrule
\multicolumn{5}{l}{\textbf{Temp}} \\
MLVU & PlotQA & 44.00 & 48.00 & \dposplain{+4.00} \\
MLVU & Order & 25.71 & 30.00 & \dposplain{+4.29} \\
MLVU & SportsQA & 36.11 & 41.67 & \dposplain{+5.56} \\
VideoMME & Temporal Perception & 67.27 & 74.55 & \dposplain{+7.27} \\
MVBench & Moving Count & 53.00 & 63.00 & \textbf{\dposplain{+10.00}} \\
\midrule
\multicolumn{5}{l}{\textbf{Temp+}} \\
MVBench & Character Order & 67.50 & 74.50 & \dposplain{+7.00} \\
MLVU & Order & 25.71 & 32.86 & \dposplain{+7.14} \\
MVBench & Counterfactual Inference & 59.50 & 68.00 & \dposplain{+8.50} \\
MLVU & SportsQA & 36.11 & 47.22 & \dposplain{+11.11} \\
VideoMME & Temporal Perception & 67.27 & 85.45 & \textbf{\dposplain{+18.18}} \\
\bottomrule
\end{tabularx}
\end{table}

\begin{figure}[t]
  \centering
  \includegraphics[width=\linewidth]{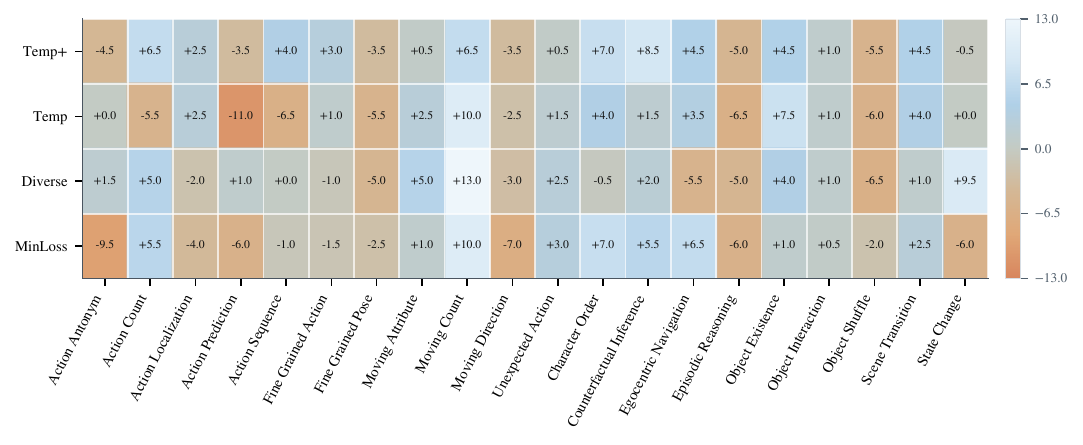}
  \caption{\textbf{MVBench Subtask Delta Heatmap.} Each cell shows the final displayed difference between a \gdo profile and the frozen benchmark-specific \uni reference on the same MVBench subtask. The heatmap displays gains, unchanged cells, and adverse outcomes across the full subtask map.}
  \label{fig:mvbench-subtask-heatmap}
\end{figure}

Figure~\ref{fig:supp-subtask-trajectories} complements the final subtask deltas in the main paper by showing how representative subtasks evolve across checkpoints.

\begin{center}
  \includegraphics[width=\textwidth]{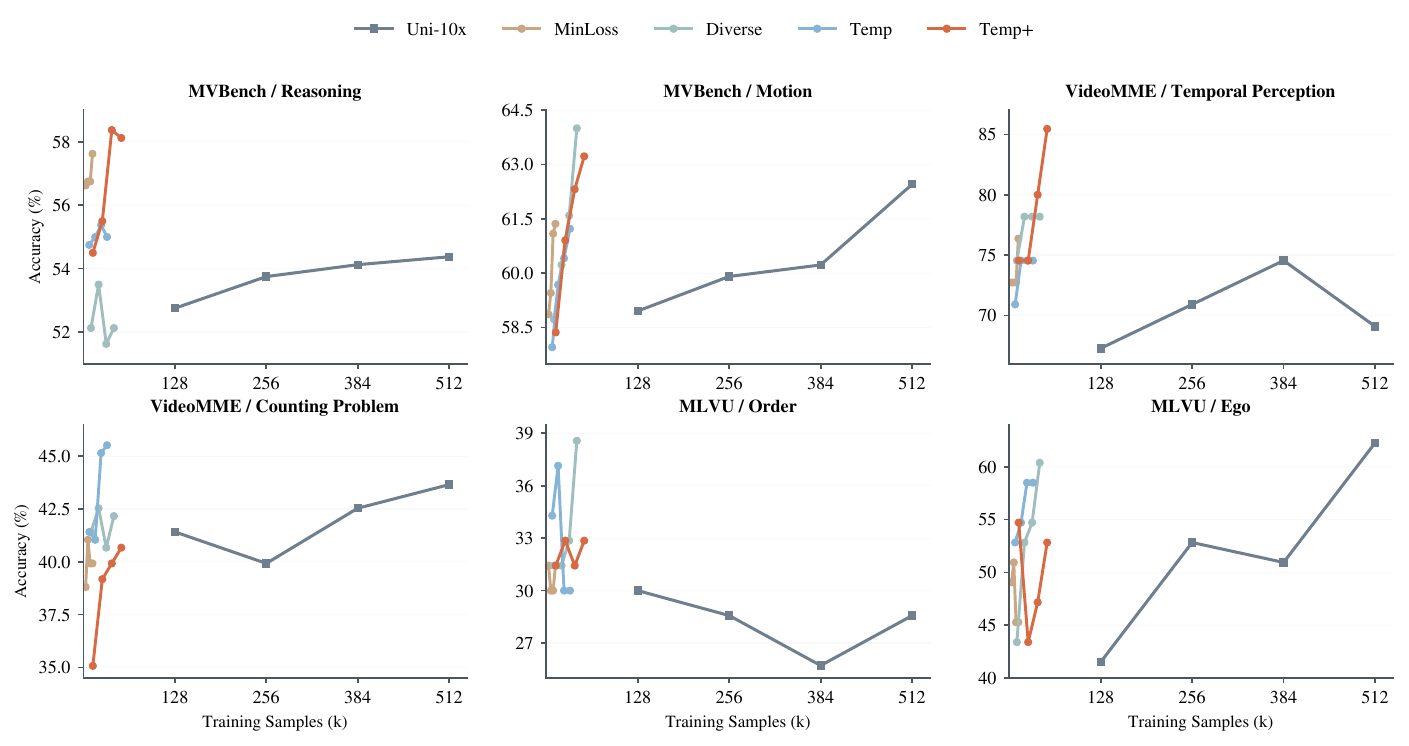}
  \captionof{figure}{\textbf{Subtask Trajectories.}
  Panels show persistent temporal gains and adverse subtasks against the frozen benchmark-specific \uni reference.}
  \label{fig:supp-subtask-trajectories}
\end{center}
\FloatBarrier

The trajectory view provides additional context for the final subtask table. VideoMME Temporal Perception and MLVU Order separate progressively as temporal pressure increases, while MVBench Reasoning shows that the gain is not restricted to one notion of motion. These trajectories show that several benchmark-level differences persist across recorded checkpoints and concentrate in identifiable temporal and reasoning subtasks. The adverse panels are equally informative. VideoMME Counting Problem and MLVU Ego do not follow the same progression, and at some checkpoints the \uni control remains stronger. These cases show that the aggregate gains do not correspond to uniform improvements across all subtasks.

\section{Method Details}
\label{sec:supp-eq13}

\subsection{Shared Score}
\label{app:eq13-details}

\begin{table}[t]
\centering
\caption{\textbf{Temp+ Ablation.} The first row is the full Temp+ builder. Each subsequent row removes one or more score contributions; parentheses report the pp change relative to full Temp+.}
\label{tab:ablation}
\small
\begin{tabular}{lccc}
\toprule
Ablation & MVBench & VideoMME & MLVU \\
\midrule
Temp+ & 63.65 & 62.89 & 46.89 \\
Temp+ w/o VDS & 63.12 \dneg{-0.53} & 61.78 \dneg{-1.11} & 47.60 \dpos{+0.71} \\
Temp+ w/o PPL & 63.30 \dneg{-0.35} & 61.81 \dneg{-1.07} & 47.00 \dpos{+0.11} \\
Temp+ w/o SC & 62.25 \dneg{-1.40} & 62.15 \dneg{-0.74} & 43.73 \dneg{-3.16} \\
Temp+ w/o VDS/PPL/SC & 61.65 \dneg{-2.00} & 62.04 \dneg{-0.85} & 47.06 \dpos{+0.17} \\
\bottomrule
\end{tabular}
\end{table}

The released builder separates preference from admissibility. A shared score ranks candidates, and the goal profile determines which ranked candidates remain feasible under a given allocation target.

After descriptor extraction and merge, each sample keeps the descriptor vector
\[
\big[m_{\mathrm{flow}},\;
m_{\mathrm{vds}},\;
m_{\mathrm{tnc}},\;
m_{\mathrm{sc}},\;
m_{\mathrm{ppl}},\;
m_{\mathrm{cov}}\big],
\]
and the merge stage also produces one scalar \texttt{quality\_score}. The scorer then applies one ranking function to short-video samples and another to image-QA samples.

For short-video samples, the scorer uses
\begin{align}
\rho_{\mathrm{vid}}(x)
&= 0.35\,\tanh\!\left(\frac{b_{\mathrm{vid}}(x)}{3}\right)
   + 0.95\,z_{\mathrm{vds3}}(x)
   + 0.35\,z_{\mathrm{qual}}(x),
\\
b_{\mathrm{vid}}(x)
&= q_{\mathrm{text}}(x)
 + 0.85\,d(x)
 + 0.90\,a(x)
 + 0.55\,t(x)
 + 0.15\,r_{\mathrm{src}}(x).
\end{align}
$q_{\mathrm{text}}$ is a heuristic question/answer quality prior, $d$ is a medium-difficulty preference, $a$ is a bin-alignment term over duration bucket, temporal bucket, question form, question length, answer length, and source type, $t$ is the short-video temporal bonus, $r_{\mathrm{src}}$ is a source-rarity prior, $z_{\mathrm{vds3}}$ is the normalized video-dependence term, and $z_{\mathrm{qual}}$ is the normalized merged \texttt{quality\_score}. For image QA, the builder uses
\begin{align}
\rho_{\mathrm{img}}(x)
&= 0.90\,\tanh\!\left(\frac{b_{\mathrm{img}}(x)}{3}\right)
   + 0.15\,z_{\mathrm{qual}}(x),
\\
b_{\mathrm{img}}(x)
&= 1.10\,q_{\mathrm{text}}(x)
 + 0.85\,d(x)
 + 0.90\,a(x)
 + 0.15\,r_{\mathrm{src}}(x).
\end{align}
The image score omits video-only terms such as the temporal bonus and the VDS3 addition. Thus, the short-video score combines a heuristic base term, a normalized video-dependence term, and the merged quality term, while the image score follows the same logic without the video-only additions. Because the terms are normalized before mixing, the coefficients are fixed relative weights rather than incomparable raw scales. All profiles use the same scorer; only the goal-specific feasibility controls change. The profile comparison therefore isolates changes in data allocation from changes to the scoring function.

\subsection{Goal Profiles}

Table~\ref{tab:appendix-policy-knobs} makes the four released goal profiles explicit. The shared score is fixed, while the budget and feasibility controls change across profiles.

\begin{table}[h]
\centering
\small
\setlength{\tabcolsep}{6pt}
\caption{\textbf{Goal Profiles.} $N_g$ is the final 1$\times$ subset size, $r_v$ is the selected video ratio, and the remaining columns are the main feasibility controls.}
\label{tab:appendix-policy-knobs}
\begin{tabular}{lcccccc}
\toprule
Goal & $N_g$ & $r_v$ & VDS$+$ tgt & $r_v^{\min}$ & $r_v^{\max}$ & $r_{t|v}^{\min}$ \\
\midrule
MinLoss & 12.9k & 0.32 & 2600 & 0.15 & 0.32 & 0.05 \\
Diverse & 42.9k & 0.45 & 5000 & 0.25 & 0.45 & 0.15 \\
Temp    & 33.3k & 0.50 & 6500 & 0.35 & 0.50 & 0.20 \\
Temp+   & 53.3k & 0.59 & 9000 & 0.50 & 0.64 & 0.38 \\
\bottomrule
\end{tabular}
\end{table}

$N_g$ denotes the final 1$\times$ subset size used for SFT, $r_v$ the selected video ratio, VDS$+$ tgt the target count of VDS-positive samples used during budget construction, and $r_{t|v}^{\min}$ the minimum temporal-positive ratio within selected videos. MinLoss targets low-loss supervision under the smallest budget. Diverse restores broader semantic and source coverage. Temp and Temp+ raise both the selected video ratio and the temporal-positive floor. The benchmark and subtask analyses compare the outcomes associated with these preset changes under the fixed training protocol.
\FloatBarrier

\section{Limitations}
\label{sec:supp-limits}

\lvb provides the main benchmark-level limitation. It targets ultra-long videos that are weakly represented in the short-video LLaVA-Video pool, while the training mixture also contains substantial LLaVA-OneVision image QA. Its gain is therefore smaller than the gains on \mvb and \mlvu, although the scores still improve from MinLoss to Temp and Temp+. The temporal-necessity signal is also only a question-level proxy, and the released profiles vary feasibility constraints more than the shared score itself. Stronger goal-specific scorers and native ultra-long-video pools remain open directions.

\FloatBarrier
\section{Qualitative Examples}
\label{sec:supp-qual}
\captionsetup{skip=2pt}
\vspace{-0.4em}
\begin{center}
  \includegraphics[width=0.98\textwidth]{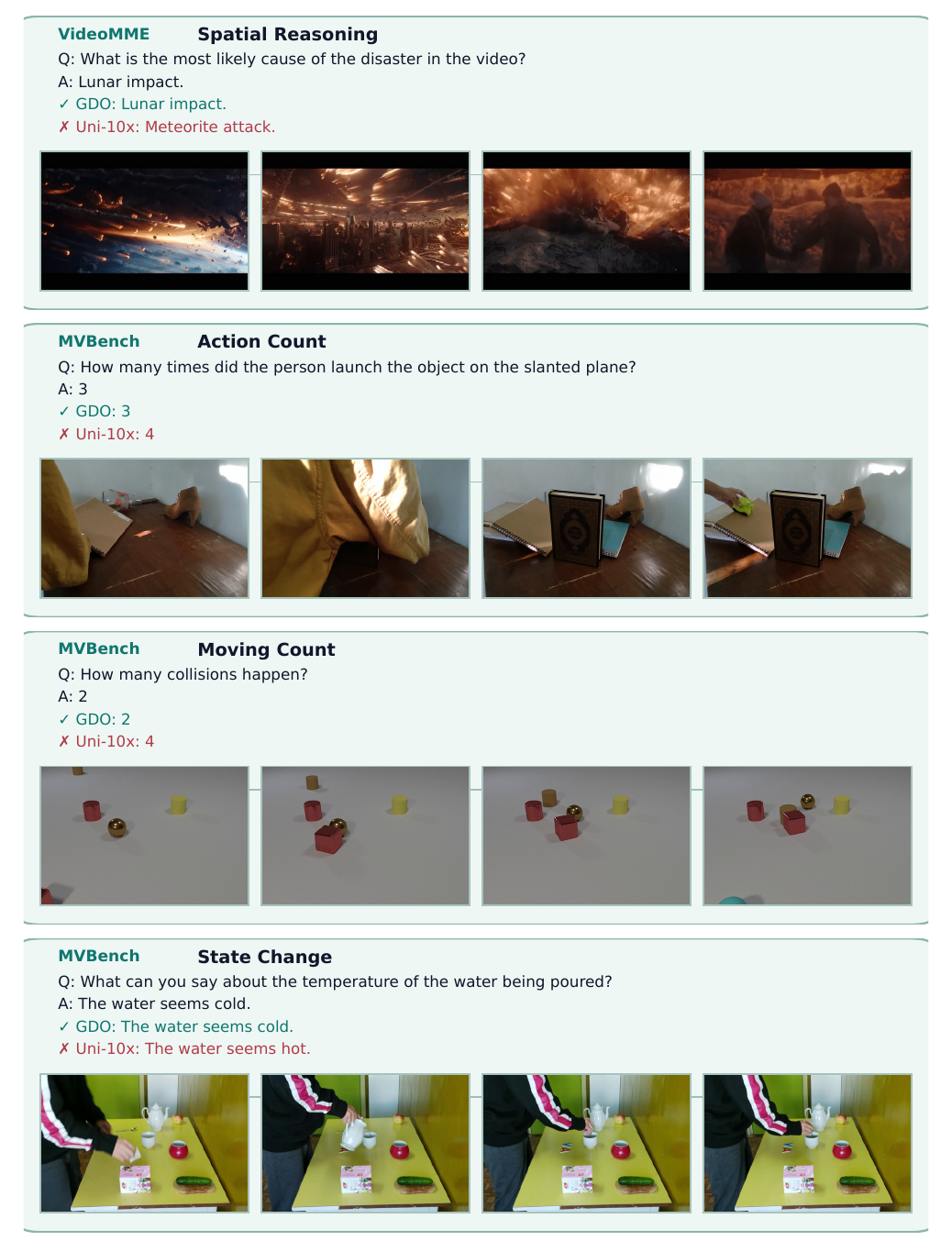}
  \captionof{figure}{\textbf{\gdo Wins.}
  Representative cases where \gdo resolves the decisive cue before \uni, including causal interpretation in \textsc{VideoMME}, repeated-action counting in \textsc{MVBench} Action Count, collision counting in Moving Count, and temperature-based state inference in State Change. Each row shows sampled frames, the question, the answer, and both model outputs.}
  \label{fig:supp-qual-win}
\end{center}
\clearpage

\begin{center}
  \includegraphics[width=0.98\textwidth]{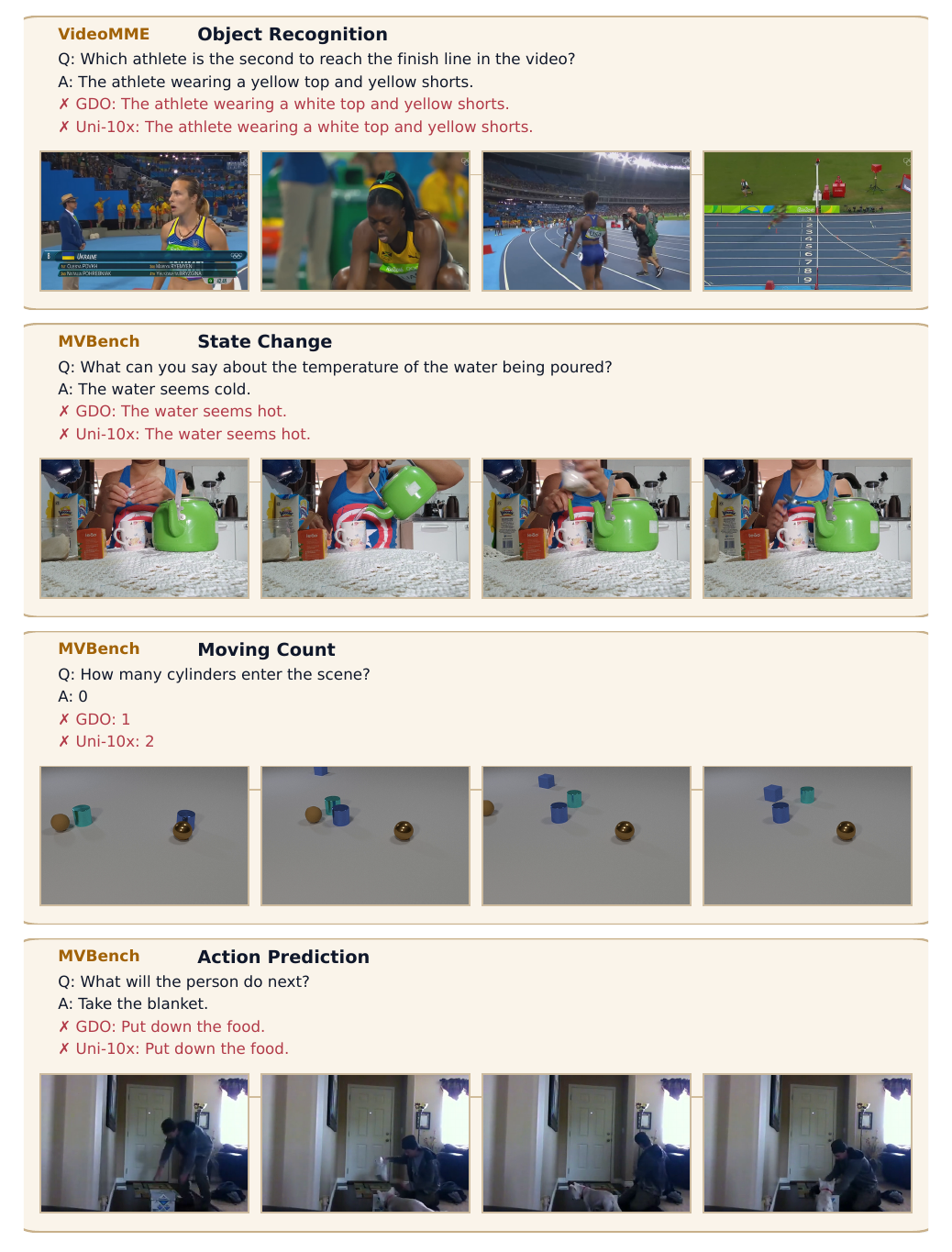}
  \captionof{figure}{\textbf{Difficult Cases for Both.}
  Cases that remain hard for both models under the fixed train/eval contract, including finish-order recognition in \textsc{VideoMME}, subtle temperature inference in State Change, small-object counting in Moving Count, and next-action prediction in Action Prediction. These rows make the remaining failure modes concrete.}
  \label{fig:supp-qual-both}
\end{center}
\clearpage

\begin{center}
  \includegraphics[width=0.98\textwidth]{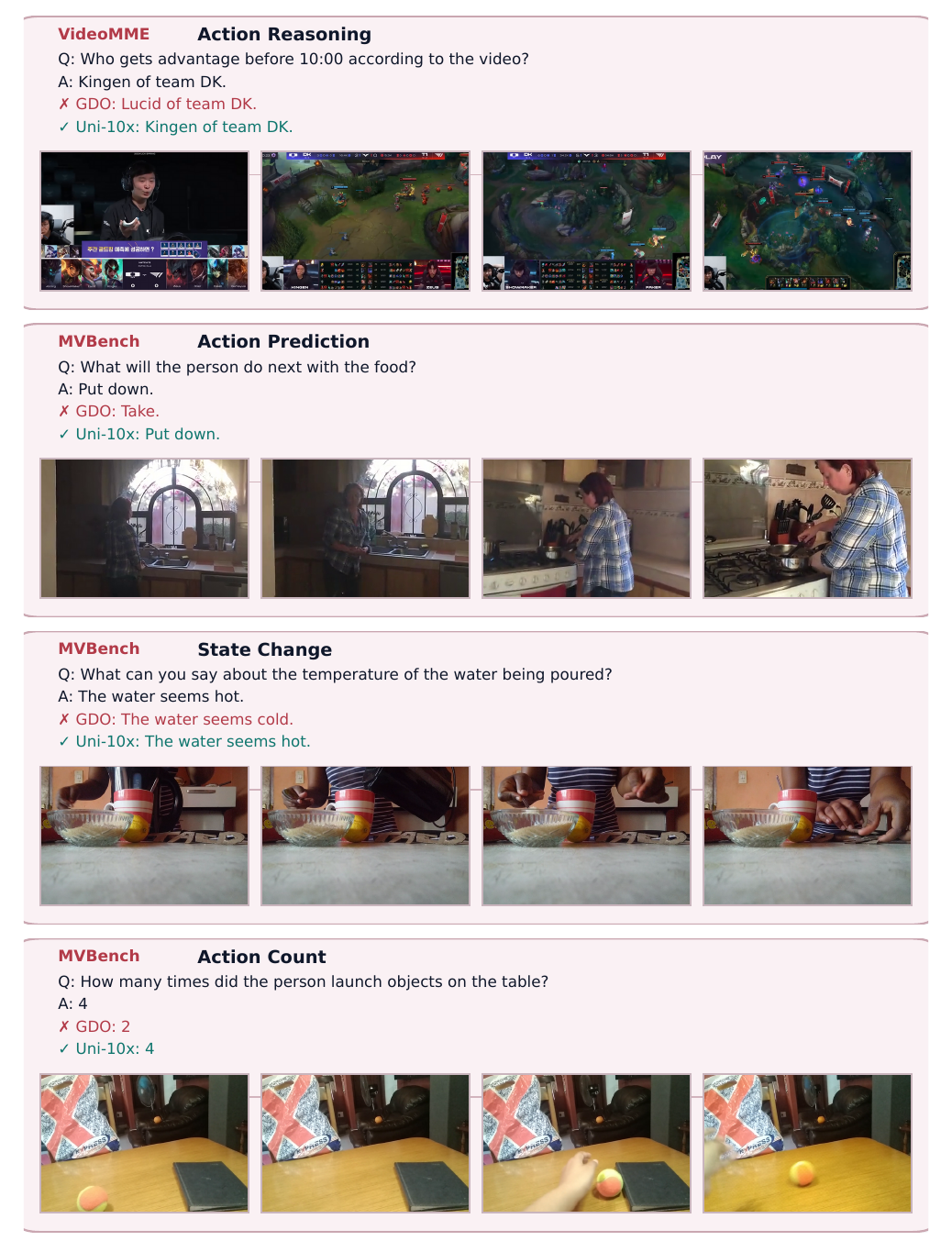}
  \captionof{figure}{\textbf{\uni Wins.}
  Cases where the frozen benchmark-specific \uni reference remains better, including esports action reasoning in \textsc{VideoMME}, object-heavy next-action prediction, subtle water-temperature inference, and repeated-action counting. These rows show that the gain from temporal allocation is not uniform across all question types.}
  \label{fig:supp-qual-uni}
\end{center}

\end{document}